\documentclass[sigconf]{acmart}
\AtBeginDocument{%
  }

\setcopyright{acmlicensed}
\copyrightyear{2025} 
\acmYear{2025} 
\setcopyright{acmlicensed}\acmConference[MM '25]{Proceedings of the 33rd
 ACM International Conference on Multimedia}{October 27--31, 2025}{Dublin,
 Ireland}
 \acmBooktitle{Proceedings of the 33rd ACM International Conference on
 Multimedia (MM '25), October 27--31, 2025, Dublin, Ireland}
 \acmDOI{10.1145/3746027.3758217}
 \acmISBN{979-8-4007-2035-2/2025/10}

\usepackage{bm}
\usepackage{multirow}
\usepackage{array}
\usepackage{multicol}
\usepackage{enumerate}
\usepackage{comment}
\usepackage{tabularx}
\usepackage{booktabs}
\usepackage{makecell}
\usepackage{graphicx}

\usepackage{utfsym}
\begin{document}
 
\title{PA-HOI: A Physics-Aware Human and Object Interaction Dataset}
 
\author{Ruiyan Wang}
\affiliation{%
  \institution{Shanghai Jiao Tong University}
  \city{Shanghai}
  \country{China}
}
\email{rywang0627@sjtu.edu.cn} 

\author{Lin Zuo}
\affiliation{%
  \institution{Shanghai Jiao Tong University}
  \city{Shanghai}
  \country{China}}
\email{zuo_lin@sjtu.edu.cn}

\author{Zonghao Lin}
\affiliation{%
  \institution{Shanghai Jiao Tong University}
  \city{Shanghai}
  \country{China}
}
\email{ruianlzh@sjtu.edu.cn}

\author{Qiang Wang}
\affiliation{%
  \institution{VisionStar Information Technology Co., Ltd.}
  \city{Shanghai}
  \country{China}
}
\email{wq@sightp.com}

\author{Zhengxue Cheng}
\authornote{Corresponding authors.}
\affiliation{%
  \institution{Shanghai Jiao Tong University}
  \city{Shanghai}
  \country{China}
}
\email{zxcheng@sjtu.edu.cn}

\author{Rong Xie}
\affiliation{%
  \institution{Shanghai Jiao Tong University}
  \city{Shanghai}
  \country{China}
}
\email{xierong@sjtu.edu.cn}

\author{Jun Ling}
\authornotemark[1] 
\affiliation{
  \institution{Shanghai Jiao Tong University}
  \city{Shanghai}
  \country{China}
}
\email{lingjun@sjtu.edu.cn}

\author{Li Song}
\authornotemark[1]  
\affiliation{
  \institution{Shanghai Jiao Tong University}
  \city{Shanghai}
  \country{China}
}
\email{song_li@sjtu.edu.cn}

\renewcommand{\shortauthors}{Ruiyan Wang et al.}


\begin{teaserfigure}
\centering
\includegraphics[width=18cm]{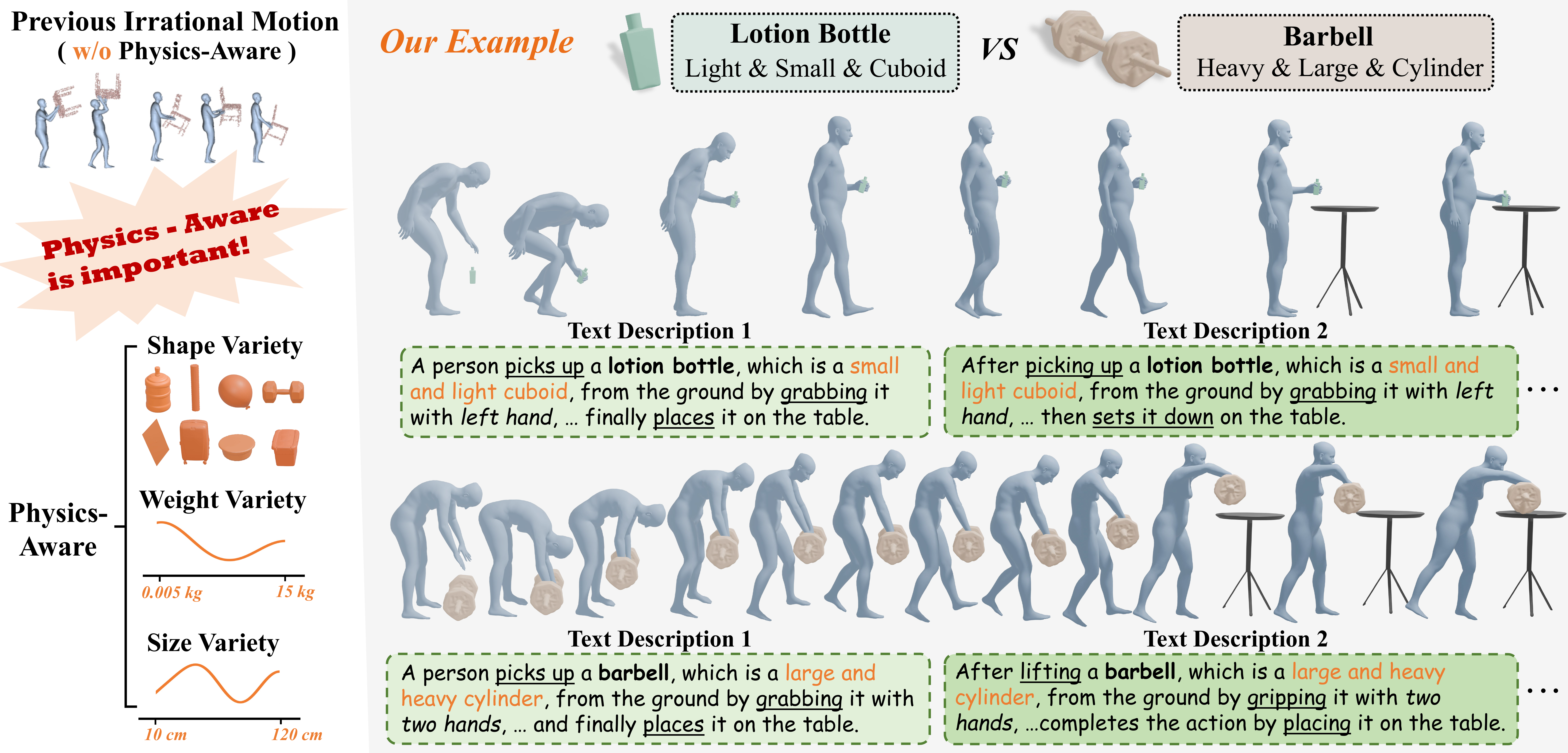}
\setlength{\abovecaptionskip}{5pt}
\caption{\textbf{Overview.} PA-HOI focuses on the impact of the physical attributes of objects on human-object interactions, with each sequence accompanied by multiple textual descriptions.}
\label{fig:teaser}
\end{teaserfigure}
\begin{abstract}
The Human-Object Interaction (HOI) task explores the dynamic interactions between humans and objects in physical environments, providing essential biomechanical and cognitive-behavioral foundations for fields such as robotics, virtual reality, and human-computer interaction. However, existing HOI data sets focus on details of affordance, often neglecting the influence of physical properties of objects on human long-term motion. To bridge this gap, we introduce the PA-HOI Motion Capture dataset, which highlights the impact of objects’ physical attributes on human motion dynamics, including human posture, moving velocity, and other motion characteristics. The dataset comprises 562 motion sequences of human-object interactions, with each sequence performed by subjects of different genders interacting with 35 3D objects that vary in size, shape, and weight. This dataset stands out by significantly extending the scope of existing ones for understanding how the physical attributes of different objects influence human posture, speed, motion scale, and interacting strategies. We further demonstrate the applicability of the PA-HOI dataset by integrating it with existing motion generation methods, validating its capacity to transfer realistic physical awareness. Our project page and dataset are available at \url{https://rayzuo123.github.io/pa-hoi-dataset/}.
\end{abstract}
\maketitle
\begin{table*}[!t]
\centering
\setlength{\abovecaptionskip}{2pt}
\caption{Dataset comparisons. ``\#'' denotes the number of objects or sequences, and the terms `single' and `multiple' refer to the number of textual descriptions associated with each interaction sequence.}
\label{tab:dataset_comparison}
\begin{tabular}{lcccccccc}  
\toprule 
\multirow{2}{*}{{Dataset}} & \multirow{2}{*}{{\#Obj}} & \multirow{2}{*}{{\#Seq}} & \multirow{2}{*}{{Motion Capture}} & \multirow{2}{*}{{Text Annotation} }  & \multicolumn{3}{c}{{Physics-Aware}} \\
\cline{6-8}
 & & & & & {Shape Variety} & {Size Variety }& {Weight Variety} \\
\hline
GRAB~\cite{taheri2020grab} & 51 & 1134 & \usym{1F5F8} &   &\usym{1F5F8} & & \\
BEHAVE~\cite{bhatnagar2022behave} & 20 & 321 &   &    &\usym{1F5F8} &\usym{1F5F8} & \\
InterCap~\cite{huang2022intercap} & 10 & 223 &   &   &\usym{1F5F8} &\usym{1F5F8} & \\
OMOMO~\cite{li2023object} & 15 & -- & \usym{1F5F8} & \usym{1F5F8} (single)  &\usym{1F5F8} & & \\
ARCTIC~\cite{fan2023arctic} & 11 & 339 &   &   &\usym{1F5F8} & & \\
HIMO~\cite{lv2024himo} & 53 & 3376 & \usym{1F5F8} & \usym{1F5F8} (single) &\usym{1F5F8} & & \\
ParaHome~\cite{kim2024parahome} & 22 & 207 & \usym{1F5F8} & \usym{1F5F8} (single) &\usym{1F5F8} &\usym{1F5F8} & \\
FORCE~\cite{zhang2024force} &  8  &  450 & \usym{1F5F8}  &    & \usym{1F5F8} &  & \usym{1F5F8} \\
\hline
\textbf{Ours} & \textbf{35} & \textbf{562} & \usym{1F5F8} & \usym{1F5F8} (multiple) & \usym{1F5F8} & \usym{1F5F8} & \usym{1F5F8} \\
\bottomrule
\end{tabular}
\end{table*}
\section{Introduction}
\label{sec:intro}

Accurately modeling and understanding Human-Object Interaction (HOI) is crucial for advancing intelligent systems toward more realistic and practical applications in VR/AR, games, and robotics.
Benefiting from rapid advancements in computer vision, sensing technologies, and deep learning, research on HOI has achieved significant breakthroughs in recent years. Several studies~\cite{taheri2020grab, fan2023arctic, li2023object, taheri2022goal, ComA, cong2025semgeomo, zeng2025chainhoi, xue2025guiding} have been developed to investigate contact preferences in object interactions. Through systematic analyses of human grasping behavior, these works reveal strong and consistent correlations between affordance and the shape of objects, offering valuable insights into the underlying mechanisms of HOI. 

However, in real-world scenarions, human manipulation frequently involves the coordinated motion of multiple objects, which requires precise spatial positioning and motion modeling.
Therefore, several datasets~\cite{savva2016pigraphs, lv2024himo, kim2024parahome, zhang2024hoi, zhang2024force} focusing on multi-object interactions have been proposed to model the complex interactions between humans and multiple objects. Interaction such as cutting a lemon on a knifeboard using a knife in HIMO~\cite{lv2024himo} illustrates the intricate coordination required in multi-object tasks. However, existing datasets for HOI tasks often overlook the variability in interactive motions arising from differences in object attributes. For example, in object transportation scenarios, the heavier objects physically lead to slower human moving speed, larger motion scales (e.g., opening one's arms vs. reaching out one hand), and greater degrees of body tilt (e.g., bending down vs. standing up). Likewise, larger objects tend to result in more contact areas with the human body and require a broader range of limb movements. Additionally, variations in object shape imply different human postures, motions, and interaction patterns that should be subject to the physical law. The absence of physics-aware attributes in current datasets presents a significant barrier to understanding and reasoning about complex interactions within real-world physical environments.

To bridge this gap, we propose the novel \textbf{PA-HOI} dataset, a physics-aware HOI dataset that focuses on how the physical properties of objects influence human-object interaction behaviors. We focus on three key physical attributes—\textbf{shape}, \textbf{size}, and \textbf{weight}—as they significantly impact interaction behaviors, and establish standardized criteria for object selection based on these attributes. From commonly accessible objects, we curate a set of 35 objects that cover a broad spectrum of sizes (e.g., from a small bottle to a water dispenser), weights (e.g., from an empty plastic bottle to a barbell), and shapes (e.g., cubes, spheres, cylinders). 
Unlike previous datasets~\cite{fan2023arctic, taheri2020grab, lv2024himo, zhang2024force}, which limited subjects to constrained motion capture ranges, our dataset captures interactions across diverse motion trajectories and interaction types (e.g., grabbing, supporting, pushing, holding). And the selection of scenarios involving moving objects enables a more realistic representation of the impact of physical properties within interaction sequences. Our implementations enable a detailed recording of human motion dynamics, including velocity, amplitude, and posture during interaction. 

Given the lack of semantic focus in existing HOI datasets~\cite{taheri2020grab,li2023object, zhang2024force}, we adopt a fine-grained annotation approach based on textual descriptions.
In contrast to HIMO, we use a unified text template with annotation keywords to label diverse object movement scenarios, emphasizing the impact of object physical properties on human actions. Furthermore, we leverage large language models (LLMs) to regenerate multiple detailed textual descriptions for each sequence, enhancing the dataset’s diversity and generalizability for downstream tasks such as text-to-motion synthesis.

The introduction of the PA-HOI dataset represents a significant advancement in understanding the intrinsic relationship between object attributes and interaction dynamics, offering a new perspective for developing fine-grained and practically adaptable HOI models and paving the way for real-world applications. Our contributions can be summarized as follows:

\textbf{1) Dataset:} PA-HOI is a physical attribute-aware HOI dataset, which comprises high-quality and realistic human-object interactive motion sequences. Our dataset introduces objects with different physical properties and semantic-focused textual annotations and explicitly investigates how intrinsic object attributes influence human interaction behaviors.

\textbf{2) Benchmark:} 
We re-implement PA-HOI with existing motion generation methods and further develop an extended version with text augmentation on our dataset.

\textbf{3) Application:} Leveraging the unique merit of physics-aware object attributes for motion modeling, we conduct experiments and demonstrate the applicability and effectiveness of physical awareness in supporting human-object interaction modeling and motion generation.
\begin{figure*}[!tb]
\centering
    \setlength{\abovecaptionskip}{4pt}
    \includegraphics[width=1\linewidth]{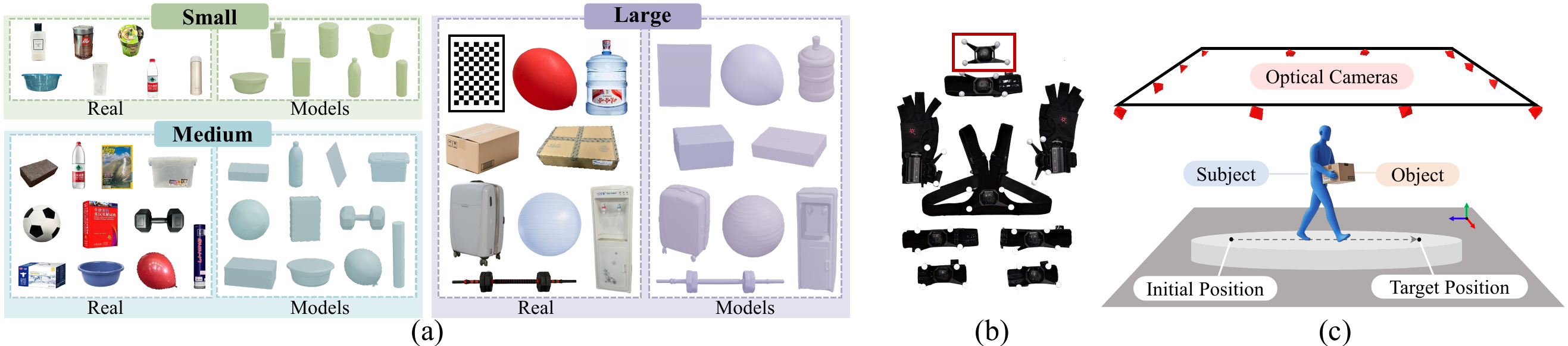}
    \caption{Overview of dataset setup. (a) The real images and renderings of 3D models for all objects. (b) The mocap (motion capture) suit is equipped with multiple hybrid tracking modules, each composed of an inertial sensor and four optical markers. (c) The capture system fuses data from optical cameras and inertial sensors to estimate human poses and object trajectories.}
\label{fig:setup}
\end{figure*} 
\section{Related Work}
\label{sec:related_work}
\textbf{Human-object interaction datasets.}
A significant line of research~\cite{taheri2020grab, hassan2021stochastic, li2023object, taheri2022goal, zhang2022couch, ComA, cong2025semgeomo, zeng2025chainhoi, xue2025guiding} in Human-Object Interaction focuses on affordance prediction, which leverages geometric cues, semantics, or learned priors to infer how and where humans can interact with objects.  
GRAB~\cite{taheri2020grab} captures authentic ``whole-body grasps'' of small 3D objects, encompassing full-body human motion, in-hand manipulation, and re-grasping actions.
OMOMO~\cite{li2023object} presents a large-scale dataset comprising high-quality human motion interactions involving large-sized objects.
Another line of research~\cite{savva2016pigraphs, lv2024himo, kim2024parahome, zhang2024hoi} addresses complex actions involving multiple objects, which introduce additional challenges, such as temporal dependencies and spatial relationships between objects.
HIMO~\cite{lv2024himo} provides a large-scale dataset capturing full-body human interactions with multiple objects, featuring accurate and diverse 4D-HOI sequences.
ParaHome~\cite{kim2024parahome} offers novel research opportunities for exploring the correlations between human motion and 3D object dynamics within realistic home environments.
While affordance-based works excel at localizing interaction regions, they may struggle to model complex actions with multiple objects. On the other hand, methods focusing on multi-object interactions can capture intricate dynamics but often require large-scale annotated datasets, which are notoriously difficult to obtain.
FORCE~\cite{zhang2024force} first focuses on the weight attribute of large objects while overlooking the shape and size variations. Moreover, the lack of text annotations limits the understanding and modeling of human interaction behaviors. 
Compared to existing datasets, our dataset provides a comprehensive consideration of multiple physical properties of objects on interaction sequences, presenting novel opportunities for research in the field of Human-Object Interaction. 

\noindent\textbf{Text-to-Motion Generation.}
The goal of text-to-motion generation is to produce human movements guided by textual descriptions. Driven by growing application demands and high-quality datasets~\cite{plappert2016kit, Guo_2022_CVPR, lin2023motion, xu2024inter, zhang2025motion}, emerging text-to-motion approaches~\cite{tevet2022human, chen2023executing, lu2023humantomato, jiang2023motiongpt, zhang2023generating, cha2024text2hoi, huang2024stablemofusion, hong2025salad, chen2025free} have been proposed. 
MDM~\cite{tevet2022human} proposes a tailored classifier-free diffusion-based generative framework for the human motion domain, whereas StableMoFusion~\cite{huang2024stablemofusion} introduces a footskate correction approach and demonstrates a strong balance between motion-text alignment and overall motion fidelity. 
In this paper, we employ our dataset with existing text-to-motion generation models~\cite{tevet2022human, huang2024stablemofusion} to validate its practical utility.
\begin{figure*}[!tb]
\centering
\setlength{\abovecaptionskip}{4pt}
\includegraphics[width=1\linewidth]{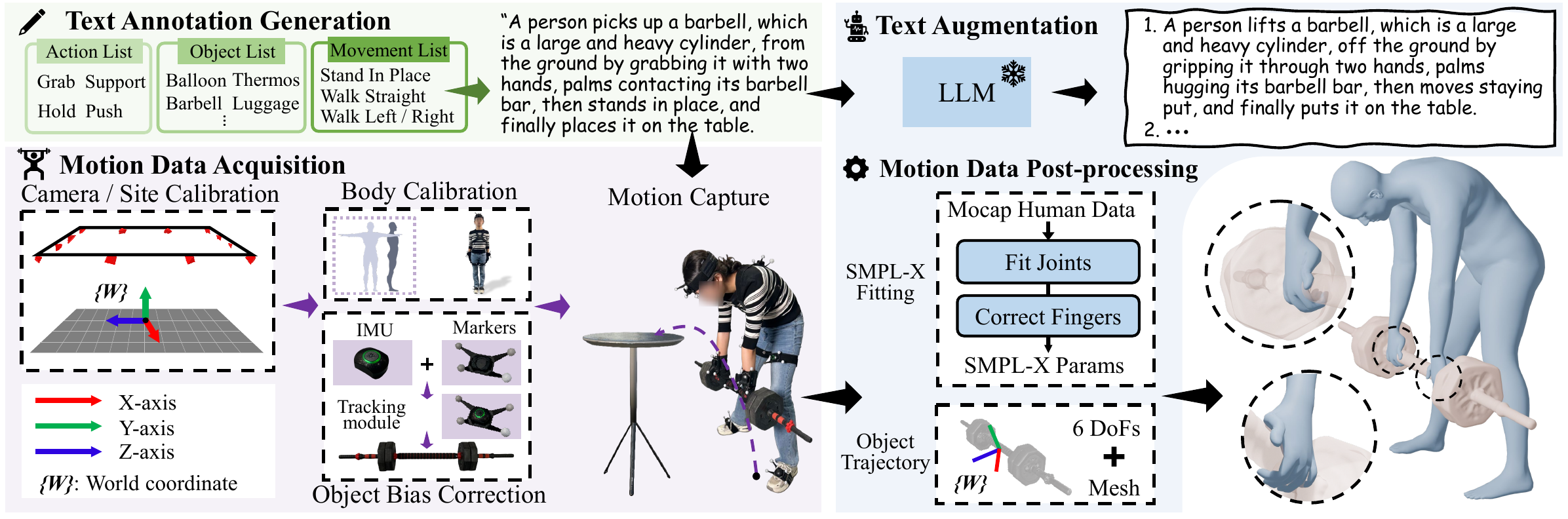}
\caption{Overview of dataset collection and post-processing. We adopt a unified textual template integrated with a predefined set of annotation keywords to describe HOI actions in Sec.~\ref{sec:Setup_text}, and employ LLMs to augment the prompt for each sequence, as detailed in Sec.~\ref{sec:Text_Augmentation}. During the motion data acquisition phase (Sec.~\ref{sec:dataset_Acquisition}), subjects first calibrate while wearing motion capture suits, and objects require inertial sensor bias correction before performing corresponding interactive actions described in the prompts. The motion data is then processed using the methods in Sec.~\ref{sec:SMPL-X_Fitting} to fit SMPL-X sequences. }
\label{fig:pipeline}
\end{figure*}
\section{Dataset Collection}
\subsection{Setup}
\subsubsection{Object Selection}
\label{sec:Setup_obj}
As physical attributes of objects influence human gestures and motions during human-object interactions, we use real-world objects with physically defined shapes for data acquisition. In particular, we select three key properties, \textit{i.e.}, size, weight, and shape, as the primary perspectives of object selection.
For simplicity, we select those that align with simple geometric primitives like spheres, cubes, and cylinders.
Generally, objects that are smaller than the human hands typically have minimal influence on hand-grasping and are therefore excluded from our scope. We classify object sizes into three categories: small, medium, and large, roughly according to the size of a palm, a forearm, and a human body, respectively. In our experiments, we define the small objects that are manipulated with one hand, medium-sized objects with either one or two hands, and large objects with two-handed interaction. 
Similarly, we categorize object weight into three groups: light, medium, and heavy. Light objects can be easily lifted with one hand, medium-weight objects may be handled with one or two hands, and heavy objects require two hands for interaction.
Based on these criteria, we selected 27 accessible everyday objects that meet the classification standards as shown in Fig.~\ref{fig:setup}(a). Among them, 8 objects can be augmented with additional weights (e.g., bottle, pot, bucket, box). To avoid ambiguity, each weighted object is treated as a distinct item, and the unweighted version is denoted with the suffix `empty'. As a result, our dataset includes a total of 35 objects.

\subsubsection{Motion Design}
\label{sec:Setup_motion}
By analyzing numerous representative behaviors in real-world human-object interaction, we define the four categories of interactions based on contact regions: `grab' refers to contact between the internal part of the hand (palm and fingers) and the object's surface; `support' means holding the object from the bottom; `hold' involves embracing the object with body parts other than the hand, such as the arm or chest; and `push' refers to propelling a large object along the ground using both hands.
`Grab' and `support' are typically associated with small and medium-sized objects, while `hold' and `push' are more commonly used with large objects.
We categorize motion movements into four types: walking straight, walking left, walking right, and interacting in place. Each motion starts from a different initial position, but all objects are eventually placed at the same target location.
The distribution of actions and movements is shown in Fig.~\ref{fig:data_analysis} (a) and (b). The size and weight of the objects result in a greater number of `grab' action sequences, while `push' and `support' actions occur less frequently.

\subsubsection{Text Annotation Generation}
\label{sec:Setup_text}
An interaction sequence involving object movement can be divided into three stages: picking up the object, moving it, and placing it down. The `picking up' and `placing down' stages are represented using two components — the `action' and the `object' — while the `moving' stage additionally includes the `movement'. The entire sequence can be expressed using the following textual template:
\begin{quote}
\ttfamily
\small
A person picks up a/an <object>, [physical attributes] (from the ground) by <action> it, then <move>, and finally places it on the table.
\end{quote}
As shown in Fig.~\ref{fig:pipeline}, we generate action annotations by replacing the keywords in this textual template with corresponding annotation terms, which are randomly selected from predefined lists of \textit{action}, \textit{object}, and \textit{movement}.

\subsubsection{Hybrid Motion Capture System}
\label{sec:Setup_system}
Traditional multi-view RGB-D systems~\cite{bhatnagar2022behave, huang2022intercap} struggle to handle occlusion by large-scale objects in interaction scenarios, often resulting in erroneous pose estimates.
To overcome this limitation, we adopt the optical motion capture approach used in~\cite{taheri2020grab, lv2024himo}, which captures high-quality human motion sequences directly, without relying on image-based motion reconstruction.
In this work, we employ the Noitom PN Hybrid VTS System~\cite{noitom}, which comprises 12 infrared cameras and a full-body optical-inertial hybrid MoCap setup, including the Noitom Perception Neuron Studio (PNS) inertial gloves, as illustrated in Fig.~\ref{fig:setup}(b).
The MoCap system is calibrated to match the skeletal dimensions of individual subjects, thereby improving the accuracy of joint data.
Additionally, the PNS gloves utilize inertial computation to enable precise tracking of finger joints, even under occlusion.
\begin{figure*}[!ht]
\centering
\setlength{\abovecaptionskip}{4pt}
\includegraphics[width=1\linewidth]{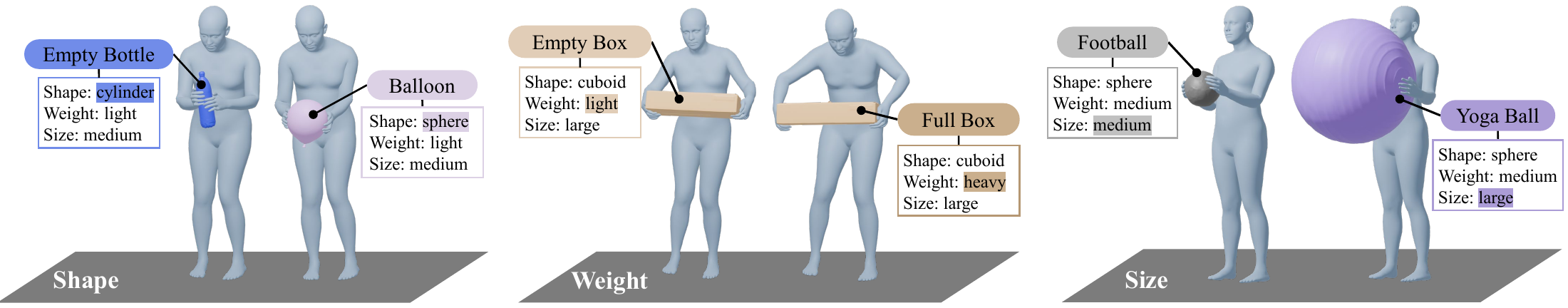}
\caption{Visualization results of the impact of object attributes on human motion during interaction. In each experiment, the other two attributes were kept consistent to isolate the effect of the specific attribute under investigation.}
\label{fig:attribute_comp}
\end{figure*}
\label{sec:dataset_postprocessing}
\begin{figure*}[!ht]
\centering
\setlength{\abovecaptionskip}{4pt}
\includegraphics[width=1\linewidth]{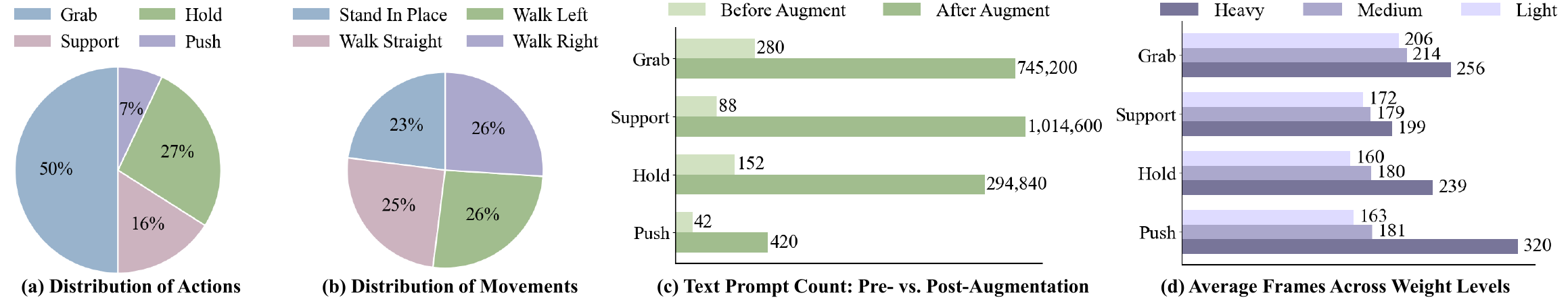}
\caption{The dataset analysis. (a) and (b) show the proportional distributions of various motions and motion paths within the interaction sequences, respectively. (c) presents the changes in the number of textual descriptions before and after augmentation with different action prompts. (d) reports the average number of frames in interaction sequences involving different actions and objects of varying weights.}
\label{fig:data_analysis}
\end{figure*}
\subsection{Motion Data Acquisition}
\label{sec:dataset_Acquisition}
\subsubsection{Objects Tracking}
We select 35 common objects from daily life, varying in size, shape, and weight, as targets for data acquisition. Initially, their 3D models are obtained using Tripo AI~\cite{tripo3d} and subsequently scaled to match real-world dimensions. In contrast to existing HOI works~\cite{taheri2020grab, lv2024himo}, which often rely on 3D-printed replicas, we use real physical objects for capturing interactions. This approach preserves the objects’ physical properties and yields highly realistic interaction motion sequences.
We use the Noitom optical-inertial hybrid system to capture the 6-DoF representations (translation and rotation) of objects during motion. As illustrated in the red box in Fig.~\ref{fig:setup}(b), each hybrid tracking module comprises inertial sensors mounted on a K-type rigid body affixed to the object’s surface, enabling real-time tracking of its 6-DoF representation.
The preprocessed 3D object models are imported into the Noitom PN Hybrid VTS software, where the initial offsets are set based on fixed reference points to achieve accurate spatial alignment. During data recording, the interactions between subjects and objects can be visualized in real time within the software.

\subsubsection{Motion Capture}
After importing the object models and completing the calibration of the motion capture equipment, subjects stand upright at the initial position in a neutral pose. Upon the start of recording, each subject performs predefined interaction actions and follows specified walking trajectories as instructed by the prompts. Each sequence concludes when the subject places the object on the designated target table and returns to an upright, stationary pose.
\section{Dataset Post-processing}
\subsection{Text Augmentation}
\label{sec:Text_Augmentation}
To reduce model overfitting caused by fixed text templates, we use LLMs to diversify motion captions. Specifically, we expand the action and movement vocabularies and introduce varied linguistic expressions. For example, for the action `grab', we include alternatives such as `grip', `seize', and `grasp'. We also enrich descriptions with adverbial phrases like `with the right hand', `palm contacting its side', `using both hands', and `palm wrapped around its side'. In addition, LLMs generate captions with diverse sentence structures while preserving the original meaning.
These strategies expand the original 562 annotations to 2,055,060, as shown in Fig.~\ref{fig:data_analysis}(c).
\subsection{SMPL-X Parameters Fitting}
\label{sec:SMPL-X_Fitting}
SMPL-X~\cite{pavlakos2019expressive} is a parametric human body model widely used for generating 3D human meshes. It has been broadly adopted in 3D human reconstruction~\cite{xiu2023econ, moon2024expressive, hu2024gaussianavatar} and motion generation~\cite{lv2024himo, liu2024emage, xu2024inter, cong2025semgeomo} due to its ability to represent both body shape and motion accurately. In particular, SMPL-X excels at capturing detailed hand motions, which is critical for human-object interaction tasks. Consequently, we adopt the SMPL-X model to represent human motion in our dataset.
The motion of the SMPL-X model is determined primarily by the rotation of individual joints and the translation of the root joint. Specifically, the pose parameters include global orientation $\mathbf{o} \in \mathbb{R}^3$, root translation $\mathbf{t} \in \mathbb{R}^3$, body pose $\boldsymbol{\theta}_b \in \mathbb{R}^{21 \times 3}$, and finger pose $\boldsymbol{\theta}_h \in \mathbb{R}^{30 \times 3}$.
We fit the captured full-body motion data to the SMPL-X pose parameters following the method used in HIMO~\cite{lv2024himo}, while the shape parameters are estimated using Hybrid-X~\cite{li2025hybrik}.

\begin{table*}[ht]
\centering
\setlength{\abovecaptionskip}{4pt}
\caption{Quantitative baseline comparisons. $\pm$ indicates 95\% confidence interval, $\rightarrow$ indicates that closer to real is better. \textbf{Bold} indicates the best results and \underline{underlined} indicates the second best results.}
\label{tab:eval_comparison}
\begin{tabular}{lccccccc}  
\toprule 
\multirow{2}{*}{Method} & \multirow{2}{*}{MM-Dist$\downarrow$} & \multirow{2}{*}{FID$\downarrow$} & \multirow{2}{*}{Diversity$\rightarrow$}  & \multicolumn{3}{c}{R-Precision$\uparrow$} \\
\cline{5-7}
 & & & & top1 & top2& top3 \\
\midrule
 Real & 5.7195\textsuperscript{$\pm$0.0348} & 0.0055\textsuperscript{$\pm$0.0013}& 4.9061\textsuperscript{$\pm$0.1368} & 0.0492\textsuperscript{$\pm$0.0095} & 0.0852\textsuperscript{$\pm$0.0116}& 0.1172\textsuperscript{$\pm$0.0107} \\
\Xhline{0.1pt}

MDM~\cite{tevet2022human} & 8.3256\textsuperscript{$\pm$0.1001} & 36.1416\textsuperscript{$\pm$1.3521} & 5.9925\textsuperscript{$\pm$0.2208} & 0.0369\textsuperscript{$\pm$0.0069} & 0.0594\textsuperscript{$\pm$0.0093}& 0.0945\textsuperscript{$\pm$0.0096} \\
 
MDM* & \underline{5.9335\textsuperscript{$\pm$0.0622}}&
\underline{1.5246\textsuperscript{$\pm$0.0936}} &  \underline{4.4014\textsuperscript{$\pm$0.1709}}  & 0.0406\textsuperscript{$\pm$0.0055} & 0.0844\textsuperscript{$\pm$0.0090} & 0.1133\textsuperscript{$\pm$0.0114}\\
 
StableMoFusion~\cite{huang2024stablemofusion} & 6.2047\textsuperscript{$\pm$0.0923}  & 21.0413\textsuperscript{$\pm$1.0749}  & 3.8396\textsuperscript{$\pm$0.2652}  & \underline{0.0636\textsuperscript{$\pm$0.0090}} & \underline{0.1217\textsuperscript{$\pm$0.0208}} & \textbf{0.1864\textsuperscript{$\pm$0.043}}\\
 
StableMoFusion* &  \textbf{5.5856\textsuperscript{$\pm$0.0625}} & \textbf{0.5597\textsuperscript{$\pm$0.0111}}  &  \textbf{5.0852\textsuperscript{$\pm$0.2259}}  & \textbf{0.0692\textsuperscript{$\pm$0.0108}} & \textbf{0.1228\textsuperscript{$\pm$0.0111}}& \underline{0.1741\textsuperscript{$\pm$0.0163}}\\
\bottomrule
\end{tabular}
\end{table*}
\begin{figure}[!tb]
\centering
    \includegraphics[width=1\linewidth]{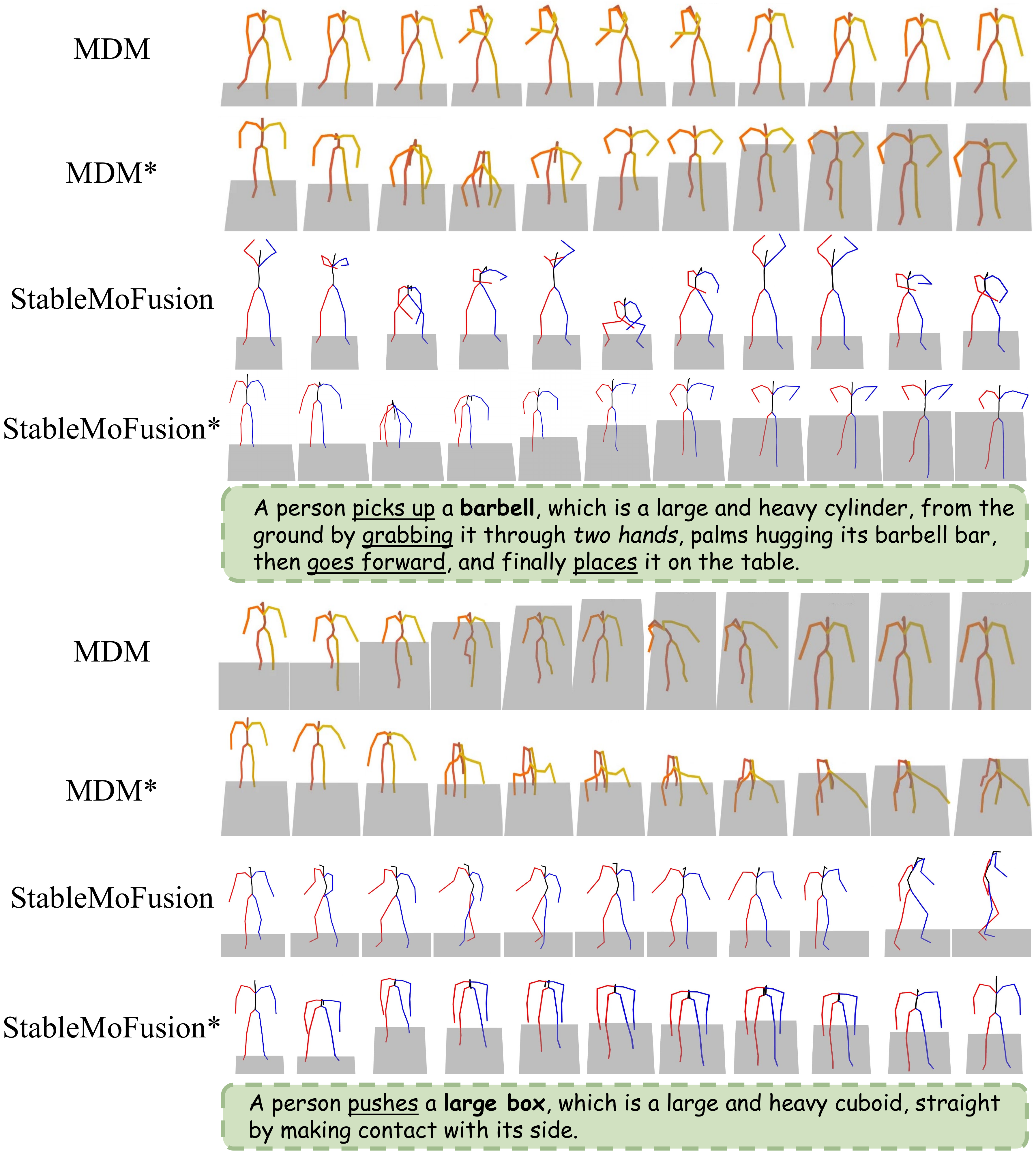}
    \caption{The qualitative comparisons. More video demos are available on our project page.}
\label{fig:comp}
\end{figure}
 
We utilize a fitting constraint $\mathcal{L}_{joint}$ on the joint positions, which can be formulated as:
\begin{equation}
\mathcal{L}_{joint} = \sum_{n=0}^{N} \sum_{j=0}^{J} \left\| P_n^j - \hat{P}_n^j \right\|_2^2,
\end{equation}
where $N$ is the frame number of the sequence, $J$ is the number of human joints, $P_n^j$ and $\hat{P}_n^j$ represent our acquired joint position and the fitting SMPL-X joint position, respectively. 
The smoothing loss $\mathcal{L}_{smooth}$ is employed to ensure temporal coherence across frames and reduce pose instability:
\begin{equation}
\mathcal{L}_{smooth} = \sum_{n=0}^{N-1} \sum_{j=0}^{J} \left\| \hat{P}_{n+1}^j - \hat{P}_n^j \right\|_2^2.
\end{equation}
The regularization term $\mathcal{L}_{reg}$ is used to regularize the SMPL-X pose parameters from deviating as:
\begin{equation}
\mathcal{L}_{reg} = \left\| \boldsymbol{\theta}_b \right\|_2^2 + \left\| \boldsymbol{\theta}_h \right\|_2^2.
\end{equation}
The overall optimization objective can be formulated as follows:
\begin{equation}
\mathcal{L} = \lambda_{j} \mathcal{L}_{joint} + \lambda_s \mathcal{L}_{smooth} + \lambda_{r} \mathcal{L}_{reg},
\end{equation}
where $\lambda_{j} = 1$, $\lambda_s = 0.1$, $\lambda_{r} = 0.01$.

\subsection{Physical Awareness}
We visualize the interactions of three object pairs performing the same action to demonstrate the influence of shape, weight, and size in Fig.~\ref {fig:attribute_comp}.
The visualization results indicate that object shape affects grasping patterns and the degree of hand flexion; an increase in weight amplifies the body bending degree of the human; and object size influences the range of motion and movement coordination.
Beyond the visualized action results, the data analysis in Fig.~\ref{fig:data_analysis}(d) reveals that heavy-weight objects require significantly more frames than light-weight ones for the same action.
Moreover, the impact of object weight on interaction duration is more pronounced for actions that are adapted to larger-sized objects, such as `hold' and `push'.
The results clearly illustrate the significant differences in how these object attributes affect human actions, underscoring the necessity of our proposed dataset for understanding and modeling diverse human-object interactions in realistic settings.

\section{Benchmark}
\label{sec:Benchmark}
In this section, we apply our dataset to existing motion synthesis methods and evaluate the generated sequences to validate the dataset’s usability and the transferability of physical awareness.

\noindent\textbf{Baseline Methods.} 
Since it's challenging for HOI methods to extract weight information from object trajectories and meshes, we instead use object attributes as text prompts to generate human motions. We adopt the prominent text-to-motion methods MDM~\cite{tevet2022human} and the recent StableMoFusion~\cite{huang2024stablemofusion} as our baselines. We re-implement our dataset to these methods and refer to the resulting models as MDM* and StableMoFusion*.

\noindent\textbf{Quantitative Comparisons.}
As shown in Tab.~\ref{tab:eval_comparison}, MDM* and StableMoFusion* achieve better performance than their baseline models, especially MM-Dist and FID. 
The results further validate the utility of our dataset by demonstrating its effectiveness in enhancing the physical plausibility of motion generation methods.

\noindent\textbf{Qualitative Results.} 
Figure.~\ref{fig:comp} shows the generation results of different models under the same prompt. Given the prompt `push a large and heavy box', the motion generated by MDM* and StableMoFusion* exhibits correct poses and greater bending, effectively reflecting the physical properties of the object. The results demonstrate that our dataset enables existing text-to-motion methods to generate motions that better conform to real-world physical constraints in HOI scenarios.  

\section{Conclusion}
\label{sec:conclusion}
In this paper, we propose the PA-HOI dataset, which delves into exploring how object physical attributes affect human motion in HOI tasks.
The dataset contains a large number of high-quality HOI sequences, covering objects with diverse sizes, weights, and shapes. Each HOI sequence is accompanied by detailed textual annotations, including precise hand positions. We further apply LLM-based augmentation to enhance generalizability across various downstream tasks.
Experiments demonstrate that the PA-HOI dataset contributes to generating HOI motions with realistic physical responses, offering new perspectives for understanding human-object interactions in fields such as robotics and virtual reality.
 
\noindent\textbf{Limitation.} Our dataset has the following limitations: \textbf{1) Facial expression:} We ignore the subjects' expressions, focusing on the impacts of various physical properties on interactive gestures and human motions, such as posture and the degree of finger flexion.
\textbf{2) Subjects:} Our dataset involves a limited number of participants. In future work, we plan to significantly increase the number of subjects to enhance the diversity and generalizability of the study.

\begin{acks}
This work was partly supported by the NSFC62431015, Science and Technology Commission of Shanghai Municipality No.24511106200, the Fundamental Research Funds for the Central Universities, Shanghai Key Laboratory of Digital Media Processing and Transmission under Grant 22DZ2229005, 111 project BP0719010.
\end{acks}

\bibliographystyle{ACM-Reference-Format}
\balance
\bibliography{sample-base}


\begin{thebibliography}{39}


\ifx \showCODEN    \undefined \def \showCODEN     #1{\unskip}     \fi
\ifx \showISBNx    \undefined \def \showISBNx     #1{\unskip}     \fi
\ifx \showISBNxiii \undefined \def \showISBNxiii  #1{\unskip}     \fi
\ifx \showISSN     \undefined \def \showISSN      #1{\unskip}     \fi
\ifx \showLCCN     \undefined \def \showLCCN      #1{\unskip}     \fi
\ifx \shownote     \undefined \def \shownote      #1{#1}          \fi
\ifx \showarticletitle \undefined \def \showarticletitle #1{#1}   \fi
\ifx \showURL      \undefined \def \showURL       {\relax}        \fi
\providecommand\bibfield[2]{#2}
\providecommand\bibinfo[2]{#2}
\providecommand\natexlab[1]{#1}
\providecommand\showeprint[2][]{arXiv:#2}

\bibitem[Bhatnagar et~al\mbox{.}(2022)]%
        {bhatnagar2022behave}
\bibfield{author}{\bibinfo{person}{Bharat~Lal Bhatnagar}, \bibinfo{person}{Xianghui Xie}, \bibinfo{person}{Ilya~A Petrov}, \bibinfo{person}{Cristian Sminchisescu}, \bibinfo{person}{Christian Theobalt}, {and} \bibinfo{person}{Gerard Pons-Moll}.} \bibinfo{year}{2022}\natexlab{}.
\newblock \showarticletitle{BEHAVE: Dataset and method for tracking human object interactions}. In \bibinfo{booktitle}{\emph{Proceedings of the IEEE/CVF Conference on Computer Vision and Pattern Recognition}}. \bibinfo{pages}{15935--15946}.
\newblock


\bibitem[Cha et~al\mbox{.}(2024)]%
        {cha2024text2hoi}
\bibfield{author}{\bibinfo{person}{Junuk Cha}, \bibinfo{person}{Jihyeon Kim}, \bibinfo{person}{Jae~Shin Yoon}, {and} \bibinfo{person}{Seungryul Baek}.} \bibinfo{year}{2024}\natexlab{}.
\newblock \showarticletitle{Text2hoi: Text-guided 3d motion generation for hand-object interaction}. In \bibinfo{booktitle}{\emph{Proceedings of the IEEE/CVF Conference on Computer Vision and Pattern Recognition}}. \bibinfo{pages}{1577--1585}.
\newblock


\bibitem[Chen et~al\mbox{.}(2025)]%
        {chen2025free}
\bibfield{author}{\bibinfo{person}{Wenshuo Chen}, \bibinfo{person}{Haozhe Jia}, \bibinfo{person}{Songning Lai}, \bibinfo{person}{Keming Wu}, \bibinfo{person}{Hongru Xiao}, \bibinfo{person}{Lijie Hu}, {and} \bibinfo{person}{Yutao Yue}.} \bibinfo{year}{2025}\natexlab{}.
\newblock \showarticletitle{Free-T2M: Frequency Enhanced Text-to-Motion Diffusion Model With Consistency Loss}.
\newblock \bibinfo{journal}{\emph{arXiv preprint arXiv:2501.18232}} (\bibinfo{year}{2025}).
\newblock


\bibitem[Chen et~al\mbox{.}(2023)]%
        {chen2023executing}
\bibfield{author}{\bibinfo{person}{Xin Chen}, \bibinfo{person}{Biao Jiang}, \bibinfo{person}{Wen Liu}, \bibinfo{person}{Zilong Huang}, \bibinfo{person}{Bin Fu}, \bibinfo{person}{Tao Chen}, {and} \bibinfo{person}{Gang Yu}.} \bibinfo{year}{2023}\natexlab{}.
\newblock \showarticletitle{Executing your Commands via Motion Diffusion in Latent Space}. In \bibinfo{booktitle}{\emph{Proceedings of the IEEE/CVF Conference on Computer Vision and Pattern Recognition}}. \bibinfo{pages}{18000--18010}.
\newblock


\bibitem[Cong et~al\mbox{.}(2025)]%
        {cong2025semgeomo}
\bibfield{author}{\bibinfo{person}{Peishan Cong}, \bibinfo{person}{Ziyi Wang}, \bibinfo{person}{Yuexin Ma}, {and} \bibinfo{person}{Xiangyu Yue}.} \bibinfo{year}{2025}\natexlab{}.
\newblock \showarticletitle{SemGeoMo: Dynamic Contextual Human Motion Generation with Semantic and Geometric Guidance}.
\newblock \bibinfo{journal}{\emph{arXiv preprint arXiv:2503.01291}} (\bibinfo{year}{2025}).
\newblock


\bibitem[Fan et~al\mbox{.}(2023)]%
        {fan2023arctic}
\bibfield{author}{\bibinfo{person}{Zicong Fan}, \bibinfo{person}{Omid Taheri}, \bibinfo{person}{Dimitrios Tzionas}, \bibinfo{person}{Muhammed Kocabas}, \bibinfo{person}{Manuel Kaufmann}, \bibinfo{person}{Michael~J Black}, {and} \bibinfo{person}{Otmar Hilliges}.} \bibinfo{year}{2023}\natexlab{}.
\newblock \showarticletitle{ARCTIC: A dataset for dexterous bimanual hand-object manipulation}. In \bibinfo{booktitle}{\emph{Proceedings of the IEEE/CVF Conference on Computer Vision and Pattern Recognition}}. \bibinfo{pages}{12943--12954}.
\newblock


\bibitem[Guo et~al\mbox{.}(2022)]%
        {Guo_2022_CVPR}
\bibfield{author}{\bibinfo{person}{Chuan Guo}, \bibinfo{person}{Shihao Zou}, \bibinfo{person}{Xinxin Zuo}, \bibinfo{person}{Sen Wang}, \bibinfo{person}{Wei Ji}, \bibinfo{person}{Xingyu Li}, {and} \bibinfo{person}{Li Cheng}.} \bibinfo{year}{2022}\natexlab{}.
\newblock \showarticletitle{Generating Diverse and Natural 3D Human Motions From Text}. In \bibinfo{booktitle}{\emph{Proceedings of the IEEE/CVF Conference on Computer Vision and Pattern Recognition (CVPR)}}. \bibinfo{pages}{5152--5161}.
\newblock


\bibitem[Hassan et~al\mbox{.}(2021)]%
        {hassan2021stochastic}
\bibfield{author}{\bibinfo{person}{Mohamed Hassan}, \bibinfo{person}{Duygu Ceylan}, \bibinfo{person}{Ruben Villegas}, \bibinfo{person}{Jun Saito}, \bibinfo{person}{Jimei Yang}, \bibinfo{person}{Yi Zhou}, {and} \bibinfo{person}{Michael~J Black}.} \bibinfo{year}{2021}\natexlab{}.
\newblock \showarticletitle{Stochastic scene-aware motion prediction}. In \bibinfo{booktitle}{\emph{Proceedings of the IEEE/CVF International Conference on Computer Vision}}. \bibinfo{pages}{11374--11384}.
\newblock


\bibitem[Hong et~al\mbox{.}(2025)]%
        {hong2025salad}
\bibfield{author}{\bibinfo{person}{Seokhyeon Hong}, \bibinfo{person}{Chaelin Kim}, \bibinfo{person}{Serin Yoon}, \bibinfo{person}{Junghyun Nam}, \bibinfo{person}{Sihun Cha}, {and} \bibinfo{person}{Junyong Noh}.} \bibinfo{year}{2025}\natexlab{}.
\newblock \showarticletitle{SALAD: Skeleton-aware Latent Diffusion for Text-driven Motion Generation and Editing}.
\newblock \bibinfo{journal}{\emph{arXiv preprint arXiv:2503.13836}} (\bibinfo{year}{2025}).
\newblock


\bibitem[Hu et~al\mbox{.}(2024)]%
        {hu2024gaussianavatar}
\bibfield{author}{\bibinfo{person}{Liangxiao Hu}, \bibinfo{person}{Hongwen Zhang}, \bibinfo{person}{Yuxiang Zhang}, \bibinfo{person}{Boyao Zhou}, \bibinfo{person}{Boning Liu}, \bibinfo{person}{Shengping Zhang}, {and} \bibinfo{person}{Liqiang Nie}.} \bibinfo{year}{2024}\natexlab{}.
\newblock \showarticletitle{Gaussianavatar: Towards realistic human avatar modeling from a single video via animatable 3d gaussians}. In \bibinfo{booktitle}{\emph{Proceedings of the IEEE/CVF conference on computer vision and pattern recognition}}. \bibinfo{pages}{634--644}.
\newblock


\bibitem[Huang et~al\mbox{.}(2022)]%
        {huang2022intercap}
\bibfield{author}{\bibinfo{person}{Yinghao Huang}, \bibinfo{person}{Omid Taheri}, \bibinfo{person}{Michael~J Black}, {and} \bibinfo{person}{Dimitrios Tzionas}.} \bibinfo{year}{2022}\natexlab{}.
\newblock \showarticletitle{InterCap: Joint markerless 3D tracking of humans and objects in interaction}. In \bibinfo{booktitle}{\emph{DAGM German Conference on Pattern Recognition}}. Springer, \bibinfo{pages}{281--299}.
\newblock


\bibitem[Huang et~al\mbox{.}(2024)]%
        {huang2024stablemofusion}
\bibfield{author}{\bibinfo{person}{Yiheng Huang}, \bibinfo{person}{Hui Yang}, \bibinfo{person}{Chuanchen Luo}, \bibinfo{person}{Yuxi Wang}, \bibinfo{person}{Shibiao Xu}, \bibinfo{person}{Zhaoxiang Zhang}, \bibinfo{person}{Man Zhang}, {and} \bibinfo{person}{Junran Peng}.} \bibinfo{year}{2024}\natexlab{}.
\newblock \showarticletitle{Stablemofusion: Towards robust and efficient diffusion-based motion generation framework}. In \bibinfo{booktitle}{\emph{Proceedings of the 32nd ACM International Conference on Multimedia}}. \bibinfo{pages}{224--232}.
\newblock


\bibitem[Jiang et~al\mbox{.}(2023)]%
        {jiang2023motiongpt}
\bibfield{author}{\bibinfo{person}{Biao Jiang}, \bibinfo{person}{Xin Chen}, \bibinfo{person}{Wen Liu}, \bibinfo{person}{Jingyi Yu}, \bibinfo{person}{Gang Yu}, {and} \bibinfo{person}{Tao Chen}.} \bibinfo{year}{2023}\natexlab{}.
\newblock \showarticletitle{Motiongpt: Human motion as a foreign language}.
\newblock \bibinfo{journal}{\emph{Advances in Neural Information Processing Systems}}  \bibinfo{volume}{36} (\bibinfo{year}{2023}), \bibinfo{pages}{20067--20079}.
\newblock


\bibitem[Kim et~al\mbox{.}(2025)]%
        {ComA}
\bibfield{author}{\bibinfo{person}{Hyeonwoo Kim}, \bibinfo{person}{Sookwan Han}, \bibinfo{person}{Patrick Kwon}, {and} \bibinfo{person}{Hanbyul Joo}.} \bibinfo{year}{2025}\natexlab{}.
\newblock \showarticletitle{Beyond the Contact: Discovering Comprehensive Affordance for 3D Objects from Pre-trained 2D Diffusion Models}. In \bibinfo{booktitle}{\emph{Computer Vision -- ECCV 2024}}, \bibfield{editor}{\bibinfo{person}{Ale{\v{s}} Leonardis}, \bibinfo{person}{Elisa Ricci}, \bibinfo{person}{Stefan Roth}, \bibinfo{person}{Olga Russakovsky}, \bibinfo{person}{Torsten Sattler}, {and} \bibinfo{person}{G{\"u}l Varol}} (Eds.). \bibinfo{publisher}{Springer Nature Switzerland}, \bibinfo{address}{Cham}, \bibinfo{pages}{400--419}.
\newblock
\showISBNx{978-3-031-72983-6}


\bibitem[Kim et~al\mbox{.}(2024)]%
        {kim2024parahome}
\bibfield{author}{\bibinfo{person}{Jeonghwan Kim}, \bibinfo{person}{Jisoo Kim}, \bibinfo{person}{Jeonghyeon Na}, {and} \bibinfo{person}{Hanbyul Joo}.} \bibinfo{year}{2024}\natexlab{}.
\newblock \showarticletitle{Parahome: Parameterizing everyday home activities towards 3d generative modeling of human-object interactions}.
\newblock \bibinfo{journal}{\emph{arXiv preprint arXiv:2401.10232}} (\bibinfo{year}{2024}).
\newblock


\bibitem[Li et~al\mbox{.}(2025)]%
        {li2025hybrik}
\bibfield{author}{\bibinfo{person}{Jiefeng Li}, \bibinfo{person}{Siyuan Bian}, \bibinfo{person}{Chao Xu}, \bibinfo{person}{Zhicun Chen}, \bibinfo{person}{Lixin Yang}, {and} \bibinfo{person}{Cewu Lu}.} \bibinfo{year}{2025}\natexlab{}.
\newblock \showarticletitle{HybrIK-X: Hybrid Analytical-Neural Inverse Kinematics for Whole-Body Mesh Recovery}.
\newblock \bibinfo{journal}{\emph{IEEE Trans. Pattern Anal. Mach. Intell.}} \bibinfo{volume}{47}, \bibinfo{number}{4} (\bibinfo{date}{Jan.} \bibinfo{year}{2025}), \bibinfo{pages}{2754–2769}.
\newblock
\showISSN{0162-8828}


\bibitem[Li et~al\mbox{.}(2023)]%
        {li2023object}
\bibfield{author}{\bibinfo{person}{Jiaman Li}, \bibinfo{person}{Jiajun Wu}, {and} \bibinfo{person}{C~Karen Liu}.} \bibinfo{year}{2023}\natexlab{}.
\newblock \showarticletitle{Object motion guided human motion synthesis}.
\newblock \bibinfo{journal}{\emph{ACM Transactions on Graphics (TOG)}} \bibinfo{volume}{42}, \bibinfo{number}{6} (\bibinfo{year}{2023}), \bibinfo{pages}{1--11}.
\newblock


\bibitem[Lin et~al\mbox{.}(2023)]%
        {lin2023motion}
\bibfield{author}{\bibinfo{person}{Jing Lin}, \bibinfo{person}{Ailing Zeng}, \bibinfo{person}{Shunlin Lu}, \bibinfo{person}{Yuanhao Cai}, \bibinfo{person}{Ruimao Zhang}, \bibinfo{person}{Haoqian Wang}, {and} \bibinfo{person}{Lei Zhang}.} \bibinfo{year}{2023}\natexlab{}.
\newblock \showarticletitle{Motion-x: A large-scale 3d expressive whole-body human motion dataset}.
\newblock \bibinfo{journal}{\emph{Advances in Neural Information Processing Systems}}  \bibinfo{volume}{36} (\bibinfo{year}{2023}), \bibinfo{pages}{25268--25280}.
\newblock


\bibitem[Liu et~al\mbox{.}(2024)]%
        {liu2024emage}
\bibfield{author}{\bibinfo{person}{Haiyang Liu}, \bibinfo{person}{Zihao Zhu}, \bibinfo{person}{Giorgio Becherini}, \bibinfo{person}{Yichen Peng}, \bibinfo{person}{Mingyang Su}, \bibinfo{person}{You Zhou}, \bibinfo{person}{Xuefei Zhe}, \bibinfo{person}{Naoya Iwamoto}, \bibinfo{person}{Bo Zheng}, {and} \bibinfo{person}{Michael~J Black}.} \bibinfo{year}{2024}\natexlab{}.
\newblock \showarticletitle{EMAGE: Towards unified holistic co-speech gesture generation via expressive masked audio gesture modeling}. In \bibinfo{booktitle}{\emph{Proceedings of the IEEE/CVF Conference on Computer Vision and Pattern Recognition}}. \bibinfo{pages}{1144--1154}.
\newblock


\bibitem[Lu et~al\mbox{.}(2023)]%
        {lu2023humantomato}
\bibfield{author}{\bibinfo{person}{Shunlin Lu}, \bibinfo{person}{Ling-Hao Chen}, \bibinfo{person}{Ailing Zeng}, \bibinfo{person}{Jing Lin}, \bibinfo{person}{Ruimao Zhang}, \bibinfo{person}{Lei Zhang}, {and} \bibinfo{person}{Heung-Yeung Shum}.} \bibinfo{year}{2023}\natexlab{}.
\newblock \showarticletitle{Humantomato: Text-aligned whole-body motion generation}.
\newblock \bibinfo{journal}{\emph{arXiv preprint arXiv:2310.12978}} (\bibinfo{year}{2023}).
\newblock


\bibitem[Lv et~al\mbox{.}(2024)]%
        {lv2024himo}
\bibfield{author}{\bibinfo{person}{Xintao Lv}, \bibinfo{person}{Liang Xu}, \bibinfo{person}{Yichao Yan}, \bibinfo{person}{Xin Jin}, \bibinfo{person}{Congsheng Xu}, \bibinfo{person}{Shuwen Wu}, \bibinfo{person}{Yifan Liu}, \bibinfo{person}{Lincheng Li}, \bibinfo{person}{Mengxiao Bi}, \bibinfo{person}{Wenjun Zeng}, {et~al\mbox{.}}} \bibinfo{year}{2024}\natexlab{}.
\newblock \showarticletitle{HIMO: A New Benchmark for Full-Body Human Interacting with Multiple Objects}. In \bibinfo{booktitle}{\emph{European Conference on Computer Vision}}. Springer, \bibinfo{pages}{300--318}.
\newblock


\bibitem[Moon et~al\mbox{.}(2024)]%
        {moon2024expressive}
\bibfield{author}{\bibinfo{person}{Gyeongsik Moon}, \bibinfo{person}{Takaaki Shiratori}, {and} \bibinfo{person}{Shunsuke Saito}.} \bibinfo{year}{2024}\natexlab{}.
\newblock \showarticletitle{Expressive whole-body 3D gaussian avatar}. In \bibinfo{booktitle}{\emph{European Conference on Computer Vision}}. Springer, \bibinfo{pages}{19--35}.
\newblock


\bibitem[Noitom({[n.\,d.]})]%
        {noitom}
\bibfield{author}{\bibinfo{person}{Noitom}.} \bibinfo{year}{[n.\,d.]}\natexlab{}.
\newblock \bibinfo{title}{Noitom PN Hybrid VTS System}.
\newblock \bibinfo{howpublished}{\url{https://noitom.com/}}.
\newblock


\bibitem[Pavlakos et~al\mbox{.}(2019)]%
        {pavlakos2019expressive}
\bibfield{author}{\bibinfo{person}{Georgios Pavlakos}, \bibinfo{person}{Vasileios Choutas}, \bibinfo{person}{Nima Ghorbani}, \bibinfo{person}{Timo Bolkart}, \bibinfo{person}{Ahmed~AA Osman}, \bibinfo{person}{Dimitrios Tzionas}, {and} \bibinfo{person}{Michael~J Black}.} \bibinfo{year}{2019}\natexlab{}.
\newblock \showarticletitle{Expressive body capture: 3d hands, face, and body from a single image}. In \bibinfo{booktitle}{\emph{Proceedings of the IEEE/CVF conference on computer vision and pattern recognition}}. \bibinfo{pages}{10975--10985}.
\newblock


\bibitem[Plappert et~al\mbox{.}(2016)]%
        {plappert2016kit}
\bibfield{author}{\bibinfo{person}{Matthias Plappert}, \bibinfo{person}{Christian Mandery}, {and} \bibinfo{person}{Tamim Asfour}.} \bibinfo{year}{2016}\natexlab{}.
\newblock \showarticletitle{The kit motion-language dataset}.
\newblock \bibinfo{journal}{\emph{Big data}} \bibinfo{volume}{4}, \bibinfo{number}{4} (\bibinfo{year}{2016}), \bibinfo{pages}{236--252}.
\newblock


\bibitem[Savva et~al\mbox{.}(2016)]%
        {savva2016pigraphs}
\bibfield{author}{\bibinfo{person}{Manolis Savva}, \bibinfo{person}{Angel~X Chang}, \bibinfo{person}{Pat Hanrahan}, \bibinfo{person}{Matthew Fisher}, {and} \bibinfo{person}{Matthias Nie{\ss}ner}.} \bibinfo{year}{2016}\natexlab{}.
\newblock \showarticletitle{Pigraphs: learning interaction snapshots from observations}.
\newblock \bibinfo{journal}{\emph{ACM Transactions On Graphics (TOG)}} \bibinfo{volume}{35}, \bibinfo{number}{4} (\bibinfo{year}{2016}), \bibinfo{pages}{1--12}.
\newblock


\bibitem[Taheri et~al\mbox{.}(2022)]%
        {taheri2022goal}
\bibfield{author}{\bibinfo{person}{Omid Taheri}, \bibinfo{person}{Vasileios Choutas}, \bibinfo{person}{Michael~J Black}, {and} \bibinfo{person}{Dimitrios Tzionas}.} \bibinfo{year}{2022}\natexlab{}.
\newblock \showarticletitle{Goal: Generating 4d whole-body motion for hand-object grasping}. In \bibinfo{booktitle}{\emph{Proceedings of the IEEE/CVF Conference on Computer Vision and Pattern Recognition}}. \bibinfo{pages}{13263--13273}.
\newblock


\bibitem[Taheri et~al\mbox{.}(2020)]%
        {taheri2020grab}
\bibfield{author}{\bibinfo{person}{Omid Taheri}, \bibinfo{person}{Nima Ghorbani}, \bibinfo{person}{Michael~J Black}, {and} \bibinfo{person}{Dimitrios Tzionas}.} \bibinfo{year}{2020}\natexlab{}.
\newblock \showarticletitle{GRAB: A dataset of whole-body human grasping of objects}. In \bibinfo{booktitle}{\emph{Computer Vision--ECCV 2020: 16th European Conference, Glasgow, UK, August 23--28, 2020, Proceedings, Part IV 16}}. Springer, \bibinfo{pages}{581--600}.
\newblock


\bibitem[Tevet et~al\mbox{.}(2022)]%
        {tevet2022human}
\bibfield{author}{\bibinfo{person}{Guy Tevet}, \bibinfo{person}{Sigal Raab}, \bibinfo{person}{Brian Gordon}, \bibinfo{person}{Yonatan Shafir}, \bibinfo{person}{Daniel Cohen-Or}, {and} \bibinfo{person}{Amit~H Bermano}.} \bibinfo{year}{2022}\natexlab{}.
\newblock \showarticletitle{Human motion diffusion model}.
\newblock \bibinfo{journal}{\emph{arXiv preprint arXiv:2209.14916}} (\bibinfo{year}{2022}).
\newblock


\bibitem[Tripo3d({[n.\,d.]})]%
        {tripo3d}
\bibfield{author}{\bibinfo{person}{Tripo3d}.} \bibinfo{year}{[n.\,d.]}\natexlab{}.
\newblock \bibinfo{title}{Generate 3D model Powered by AI in One Clip, within Seconds}.
\newblock \bibinfo{howpublished}{\url{https://www.tripo3d.ai/}}.
\newblock


\bibitem[Xiu et~al\mbox{.}(2023)]%
        {xiu2023econ}
\bibfield{author}{\bibinfo{person}{Yuliang Xiu}, \bibinfo{person}{Jinlong Yang}, \bibinfo{person}{Xu Cao}, \bibinfo{person}{Dimitrios Tzionas}, {and} \bibinfo{person}{Michael~J Black}.} \bibinfo{year}{2023}\natexlab{}.
\newblock \showarticletitle{Econ: Explicit clothed humans optimized via normal integration}. In \bibinfo{booktitle}{\emph{Proceedings of the IEEE/CVF conference on computer vision and pattern recognition}}. \bibinfo{pages}{512--523}.
\newblock


\bibitem[Xu et~al\mbox{.}(2024)]%
        {xu2024inter}
\bibfield{author}{\bibinfo{person}{Liang Xu}, \bibinfo{person}{Xintao Lv}, \bibinfo{person}{Yichao Yan}, \bibinfo{person}{Xin Jin}, \bibinfo{person}{Shuwen Wu}, \bibinfo{person}{Congsheng Xu}, \bibinfo{person}{Yifan Liu}, \bibinfo{person}{Yizhou Zhou}, \bibinfo{person}{Fengyun Rao}, \bibinfo{person}{Xingdong Sheng}, {et~al\mbox{.}}} \bibinfo{year}{2024}\natexlab{}.
\newblock \showarticletitle{Inter-x: Towards versatile human-human interaction analysis}. In \bibinfo{booktitle}{\emph{Proceedings of the IEEE/CVF Conference on Computer Vision and Pattern Recognition}}. \bibinfo{pages}{22260--22271}.
\newblock


\bibitem[Xue et~al\mbox{.}(2025)]%
        {xue2025guiding}
\bibfield{author}{\bibinfo{person}{Mengqing Xue}, \bibinfo{person}{Yifei Liu}, \bibinfo{person}{Ling Guo}, \bibinfo{person}{Shaoli Huang}, {and} \bibinfo{person}{Changxing Ding}.} \bibinfo{year}{2025}\natexlab{}.
\newblock \showarticletitle{Guiding Human-Object Interactions with Rich Geometry and Relations}.
\newblock \bibinfo{journal}{\emph{arXiv preprint arXiv:2503.20172}} (\bibinfo{year}{2025}).
\newblock


\bibitem[Zeng et~al\mbox{.}(2025)]%
        {zeng2025chainhoi}
\bibfield{author}{\bibinfo{person}{Ling-An Zeng}, \bibinfo{person}{Guohong Huang}, \bibinfo{person}{Yi-Lin Wei}, \bibinfo{person}{Shengbo Gu}, \bibinfo{person}{Yu-Ming Tang}, \bibinfo{person}{Jingke Meng}, {and} \bibinfo{person}{Wei-Shi Zheng}.} \bibinfo{year}{2025}\natexlab{}.
\newblock \showarticletitle{ChainHOI: Joint-based Kinematic Chain Modeling for Human-Object Interaction Generation}.
\newblock \bibinfo{journal}{\emph{arXiv preprint arXiv:2503.13130}} (\bibinfo{year}{2025}).
\newblock


\bibitem[Zhang et~al\mbox{.}(2024b)]%
        {zhang2024hoi}
\bibfield{author}{\bibinfo{person}{Juze Zhang}, \bibinfo{person}{Jingyan Zhang}, \bibinfo{person}{Zining Song}, \bibinfo{person}{Zhanhe Shi}, \bibinfo{person}{Chengfeng Zhao}, \bibinfo{person}{Ye Shi}, \bibinfo{person}{Jingyi Yu}, \bibinfo{person}{Lan Xu}, {and} \bibinfo{person}{Jingya Wang}.} \bibinfo{year}{2024}\natexlab{b}.
\newblock \showarticletitle{HOI-M\^{} 3: Capture Multiple Humans and Objects Interaction within Contextual Environment}. In \bibinfo{booktitle}{\emph{Proceedings of the IEEE/CVF Conference on Computer Vision and Pattern Recognition}}. \bibinfo{pages}{516--526}.
\newblock


\bibitem[Zhang et~al\mbox{.}(2023)]%
        {zhang2023generating}
\bibfield{author}{\bibinfo{person}{Jianrong Zhang}, \bibinfo{person}{Yangsong Zhang}, \bibinfo{person}{Xiaodong Cun}, \bibinfo{person}{Yong Zhang}, \bibinfo{person}{Hongwei Zhao}, \bibinfo{person}{Hongtao Lu}, \bibinfo{person}{Xi Shen}, {and} \bibinfo{person}{Ying Shan}.} \bibinfo{year}{2023}\natexlab{}.
\newblock \showarticletitle{Generating human motion from textual descriptions with discrete representations}. In \bibinfo{booktitle}{\emph{Proceedings of the IEEE/CVF conference on computer vision and pattern recognition}}. \bibinfo{pages}{14730--14740}.
\newblock


\bibitem[Zhang et~al\mbox{.}(2022)]%
        {zhang2022couch}
\bibfield{author}{\bibinfo{person}{Xiaohan Zhang}, \bibinfo{person}{Bharat~Lal Bhatnagar}, \bibinfo{person}{Sebastian Starke}, \bibinfo{person}{Vladimir Guzov}, {and} \bibinfo{person}{Gerard Pons-Moll}.} \bibinfo{year}{2022}\natexlab{}.
\newblock \showarticletitle{Couch: Towards controllable human-chair interactions}. In \bibinfo{booktitle}{\emph{European Conference on Computer Vision}}. Springer, \bibinfo{pages}{518--535}.
\newblock


\bibitem[Zhang et~al\mbox{.}(2024a)]%
        {zhang2024force}
\bibfield{author}{\bibinfo{person}{Xiaohan Zhang}, \bibinfo{person}{Bharat~Lal Bhatnagar}, \bibinfo{person}{Sebastian Starke}, \bibinfo{person}{Ilya Petrov}, \bibinfo{person}{Vladimir Guzov}, \bibinfo{person}{Helisa Dhamo}, \bibinfo{person}{Eduardo P{\'e}rez-Pellitero}, {and} \bibinfo{person}{Gerard Pons-Moll}.} \bibinfo{year}{2024}\natexlab{a}.
\newblock \showarticletitle{Force: Dataset and method for intuitive physics guided human-object interaction}.
\newblock \bibinfo{journal}{\emph{CoRR}} (\bibinfo{year}{2024}).
\newblock


\bibitem[Zhang et~al\mbox{.}(2025)]%
        {zhang2025motion}
\bibfield{author}{\bibinfo{person}{Yuhong Zhang}, \bibinfo{person}{Jing Lin}, \bibinfo{person}{Ailing Zeng}, \bibinfo{person}{Guanlin Wu}, \bibinfo{person}{Shunlin Lu}, \bibinfo{person}{Yurong Fu}, \bibinfo{person}{Yuanhao Cai}, \bibinfo{person}{Ruimao Zhang}, \bibinfo{person}{Haoqian Wang}, {and} \bibinfo{person}{Lei Zhang}.} \bibinfo{year}{2025}\natexlab{}.
\newblock \showarticletitle{Motion-X++: A Large-Scale Multimodal 3D Whole-body Human Motion Dataset}.
\newblock \bibinfo{journal}{\emph{arXiv preprint arXiv:2501.05098}} (\bibinfo{year}{2025}).
\newblock


\end{thebibliography}


\end{document}